\documentclass[dvips,???]{imsart}
\usepackage{graphicx}
\usepackage[numbers]{natbib}
\usepackage{amsmath,amsfonts,mathrsfs,bm,graphicx,subfigure}
\usepackage{fullpage}
\usepackage[small,bf]{caption}
\usepackage{algorithm}
\usepackage{algorithmic}
\usepackage{multirow}
\usepackage{color}

\newcommand{\eop}{{\hfill\vbox{\hrule height .2pt
      \hbox{\vrule width.2pt height 6pt/Users/xhx/Desktop/graphical model/SBGLasso.pdf
      \kern 4pt
      \vrule width .2pt}
      \hrule height .2pt}} \par\bigskip}
\RequirePackage[OT1]{fontenc}
\RequirePackage{amsthm,amsmath}
\RequirePackage{natbib}

\startlocaldefs

\newtheorem{theorem}{Theorem}

\newtheorem{lemma}{Lemma}

\endlocaldefs

\begin{document}
\begin{frontmatter}
\title{Split Bregman Method for Sparse Inverse Covariance Estimation with Matrix Iteration Acceleration}
\runtitle{Efficient Estimations for Sparse Graphical Models}
\begin{aug}
\author{\fnms{Gui-Bo} \snm{Ye}
\ead[label=e1]{yeg@uci.edu}}
\and
\author{\fnms{Jian-Feng} \snm{Cai}
\ead[label=e2]{cai@math.ucla.edu}}
\and
\author{\fnms{Xiaohui} \snm{Xie}
\ead[label=e3]{xhx@ics.uci.edu}}
\address{School of Information and Computer Science, University of California, Irvine, CA 92697, USA\\
\printead{e1}}
\address{Department of Mathematics, University of California, Los Angeles, CA  90095, USA\\
\printead{e2}}
\address{School of Information and Computer Science, University of California, Irvine, CA 92697, USA\\
\printead{e3}}

\end{aug}

\begin{abstract}
We consider the problem of estimating the inverse covariance matrix by maximizing the likelihood function with a penalty added to encourage  the sparsity of the resulting matrix.  We propose a new approach based on the split Bregman method to solve the regularized maximum likelihood estimation problem.  We show that our method is significantly faster than the widely used graphical lasso method, which is based on blockwise coordinate descent, on both artificial and real-world data.  More importantly, different from the graphical lasso, the split Bregman based method is much more general, and can be applied to a class of regularization terms other than the $\ell_1$ norm.
\end{abstract}
%
%
\end{frontmatter}

\section{Introduction}
Undirected graphical models provide an efficient way to describe and explain the relationships among a set of objects (or variables), and have become a popular tool to model networks of components in a variety of applications, including internet, social networks, and gene networks \citep{Lauritzen:book:1996,YL:Biometrika:2007,Newman:SIAM:2003}.  A key step involved in constructing a graphical model for a particular application is to learn the structure of the graphical model  from a set of observations,  which is often called model selection in statistics.  If we assume the observations have a multivariate Gaussian distribution with mean $\mu$ and covariance matrix $\Sigma$,  learning the structure of a graphical model can be reformulated as a covariance selection problem \citep{Dempster:Biometrics:1972},  which aims to identify nonzero entries of the inverse covariance matrix (also known as {\it concentration matrix} or {\it precision matrix}).  The idea behind this formulation is that if the $(i,j)$-entry of $\Sigma^{-1}$  is zero, then the variable $i$ and $j$ are conditionally independent, given the other variables \citep{Edwards:book:2000,PWZZ:JASA:2009}.

The principle of parsimony suggests that among the graphical models that adequately explains the data we should select the simplest. Thus, it is natural to impose an $\ell_1$ penalty for the estimation of $\Sigma^{-1}$ to promote the sparsity of the resulting graph, as has been proposed by a number of authors  \citep{MB:AS:2006,YL:Biometrika:2007,BEd:JMLR:2008,FHT:Biostatistics:2008}.

%

It has been noted that if we fit a linear regression model to each variable using the others as predictors, the regression coefficient of variable $j$ on $i$ will be, up to a positive scalar, equal to the $(i,j)$-entry of $\Sigma^{-1}$ \citep{HTF:book:2009,PWZZ:JASA:2009}.  Based on this observation, \citet{MB:AS:2006} proposed a simple approach to the structural learning problem;  they estimate a sparse graphical model by fitting a lasso regression model for each variable, treating the variable as the response and the others as predictors. The $(i,j)$-entry of $\Sigma^{-1}$ is estimated to be non-zero if either the estimated coefficient of variable $i$ on $j$ or the estimated coefficient of variable $j$ on $i$ is nonzero (alternatively, one can use an AND rule, requiring both coefficients to be nonzero). Although this approach is computationally attractive and has been shown to be able to consistently estimate $\Sigma^{-1}$ asymptotically, it does not take the intrinsic symmetry of $\Sigma^{-1}$ into account, and could result in contradictory neighborhoods and a non-positive definite concentration matrix. Furthermore, if the same penalty parameter is used for all lasso regressions, as is commonly done to reduce the number of parameters,  the approach will not be able to correctly infer graphs with skewed degree distributions. To overcome these limitations, \citet{PWZZ:JASA:2009} proposed a symmetric regression approach;  however, the derived concentration matrix can still be non-positive-definite.

A more principled approach to covariance selection is to find the $\Sigma^{-1}$ that maximizes the log-likelihood of the data with an added $\ell_1$ penalty \citep{Tib:JRSS:1996} to encourage the sparsity of the resulting graph \citep{YL:Biometrika:2007,BEd:JMLR:2008,FHT:Biostatistics:2008}. More specifically, suppose we are given $n$ samples independently drawn from a $p$-variate Gaussian distribution: $\mathbf{x}_1,\ldots,\mathbf{x}_n \sim \mathcal{N}(\mu, \Sigma)$.  Let $S$ be the empirical covariance matrix:
\begin{equation*}
S:=\frac{1}{n}\sum_{k=1}^n(\mathbf{x}_i-\mu)(\mathbf{x}_i-\mu)^T.
\end{equation*}
Denote $\Theta=\Sigma^{-1}$. The goal is to find the $\Theta^*$ that maximizes the penalized log-likelihood
\begin{equation}\label{min l1 penalized likelihood}
\log\det \Theta-\hbox{tr}(S\Theta)-\lambda\sum_{i\neq j}|\Theta_{ij}|,
\end{equation}
subject to the constraint that $\Theta$ is positive definite.  Here, we use tr to denote the trace and $\Theta_{ij}$ to denote the $(i,j)$-entry of $\Theta$. Due to the $\ell_1$ penalty term and the explicit positive definite constraint on $\Theta$, the method leads to a sparse estimation of the concentration matrix that is guaranteed to be positive definite. The simpler approach of \citet{MB:AS:2006} can be viewed as an approximation to the penalized maximum likelihood formulation \citep{BEd:JMLR:2008,FHT:Biostatistics:2008}.

The objective function in \eqref{min l1 penalized likelihood} is strictly convex, so a global optimal solution is guaranteed to exist and be unique. Finding the optimal solution can, however, be computationally challenging due to  the $\log\det$ term appeared in the likelihood function and the nondifferentiability of  the $\ell_1$ penalty. \citet{YL:Biometrika:2007} solved \eqref{min l1 penalized likelihood} using the interior-point method for the ``maxdet" problem,   which is prohibitive for problems with more than tens of variables due to its memory requirements and computational complexity. \citet{BEd:JMLR:2008} developed  a blockwise coordinate descent method to solve \eqref{min l1 penalized likelihood}, after noting that the dual of  \eqref{min l1 penalized likelihood} can be reduced to a lasso regression problem if one focuses on optimizing only one row or column of $\Omega$.  Exploiting this observation further, \citet{FHT:Biostatistics:2008} used the coordinate descent procedure to solve the lasso regression arising in the dual of the blockwise coordinate descent, and implemented an efficient method, called ``graphical lasso", to solve \eqref{min l1 penalized likelihood}.  The graphical lasso is remarkably fast; it can solves a $p=1000$ dimension ($\sim$500,000 parameters) problem within a minute. However, a major limitation of the blockwise coordinate descent methods, including the graphical lasso, is that their derivation is specific to the $\ell_1$ penalty. This limitation prevents them from being applied to other types of penalties, such as  a combination of $\ell_1$ and $\ell_2$ penalty used in  the elastic net \citep{ZH:JRSSB:2005}, the fused lasso penalty \citep{TSRZK:JRSS:2005}, and the group lasso penalty \citep{YL:JRSS:2006}, which have been found to be more useful than the simple $\ell_1$ penalty in a number of applications.

In this paper, we propose a new algorithm based on the split Bregman method \cite{GO:SIAM:2009,CaiOsherShen:MMS:2009} to solve the penalized maximum likelihood estimation problem \eqref{min l1 penalized likelihood}.  We show that our method is not only substantially faster than the graphical lasso method, but can also be easily extended to deal with a broad range of penalty terms.

Although the Bregman iteration was an old technique proposed in the sixties \citep{Bregman:Akad:1967,Cetin:SP:1989}, it gained significant interest only recently after Osher and others demonstrated its high efficiency for sparsity recovery in a wide range of problems, including image restoration   \citep{OBGXY:MMS:2005,GO:SIAM:2009,CaiOsherShen:MMS:2009}, compressed sensing \citep{COS:MC:2009,OMDY:CMS:2010,YOGD:SIAM:2008}, matrix completion \citep{CCS:SIOPT:2010}, low rank matrix recovery \citep{CLMW:Arxive:2009}, and general fused lasso problems \citep{YeXie:CSDA:2010}.  By introducing an auxiliary variable, we show that the optimization problem in \eqref{min l1 penalized likelihood} can be reformulated such that the log-likelihood term and the penalty term interact only through an equality constraint, thereby enabling an efficient application of the split Bregman method to solve the optimization problem.

The rest of the paper is organized as follows. In Section \ref{sec general SBGM}, we first present a generalized sparse inverse covariance matrix estimation problem, which includes \eqref{min l1 penalized likelihood} as a special case. We then derive a split Bregman algorithm, called SBGM, to solve the generalized problem (Subsection \ref{subsec framework}). The convergence property of the algorithm is also given.  SBGM consists of three update steps with the first update being the major time-consuming one. In Subsection \ref{subsec algorithm}, we provide an explicit formula for the first update step and propose an efficient Newton method to solve the resulting matrix quadratic equation.  In Section \ref{sec example}, we implement SBGM to solve the special case  \eqref{min l1 penalized likelihood} .  In Section \ref{sec experiment}, we illustrate the utility of our algorithm and compare its performance to the graphical lasso method using both simulated data and gene expression data.

\section{The split Bregman method for generalized sparse graphical models (SBGM)}\label{sec general SBGM}
The split Bregman method was originally proposed by Osher and coauthors to solve total variation based image restoration problems \citep{GO:SIAM:2009}. It was later found to be either equivalent or closely related to a number of other existing optimization algorithms, including Douglas-Rachford splitting \citep{WuTai:SIJISC:2010}, the alternating direction method of multipliers (ADMM) \citep{GabayMercier:CMA:1976,GlowinskiMarrocco:RFAIRO:1975,GO:SIAM:2009,Esser:CAM:2009} and the method of multipliers \citep{Rockafellar:MP:1973}. Because of its fast convergence and the easiness of implementation, it is increasingly becoming a method of choice for solving large-scale sparsity recovery problems \citep{CLMW:Arxive:2009,CaiOsherShen:MMS:2009,YeXie:CSDA:2010}.

In this section, we first extend the $\ell_1$ regularized maximum likelihood inverse covariance matrix estimation problem  \eqref{min l1 penalized likelihood}  to allow for a more general class of regularization terms. We  reformulate the problem by introducing an auxiliary variable.  We then proceed to derive a split Bregman method to solve the reformulated problem.
%

\subsection{A general problem and its reformulation}\label{subsec general problem}
We derive our algorithm in a more general setting than the one described in \eqref{min l1 penalized likelihood}. Instead of using the $\ell_1$-norm penalty, we consider a more general penalty $\phi(\Theta)$, which is only required to be convex and satisfy $\phi(\Theta)=\phi(\Theta^T)$. The $\ell_1$ norm and a number of other types, including those used in elastic net \citep{ZH:JRSSB:2005}, fused lasso \citep{TSRZK:JRSS:2005}, and group lasso \citep{YL:JRSS:2006}, can be viewed as a special case of the general penalty term. We find $\Theta^*$ by solving the following constrained optimization problem
\begin{equation}\label{min general likelihood}
\min_{\Theta\succ 0}-\log\det\Theta+\hbox{tr}(S\Theta)+\lambda\phi(\Theta),
\end{equation}
where $\Theta\succ 0$ means that $\Theta$ is positive definite, and $\lambda\in \mathbb{R}_+$ is a regularization parameter. If we choose $\phi(\Theta)=\sum_{i\neq j}|\Theta_{ij}|$, the general problem reduces to the problem \eqref{min l1 penalized likelihood}.

The log-likelihood term and the regularization term in \eqref{min general likelihood} are coupled, which makes the optimization problem difficult to solve. However, the two terms can be decoupled if we introduce an auxiliary variable to transfer the coupling from the objective function to the constraints.  More specially, the problem \eqref{min general likelihood} is equivalent to the following  problem
\begin{eqnarray}\label{min constraint general likelihood}
 &&\min-\log\det\Theta+\hbox{tr}(S\Theta)+
 \lambda\phi(A)\nonumber\\
 &&\ \hbox{s.t.}\quad A=\Theta\nonumber\\
 &&\qquad\quad \Theta\succ 0.
\end{eqnarray}

The introduction of the new variable of $A$ is a key step of our algorithm, which makes the problem amenable to a split Bregman procedure to be detailed below.

\subsection{Derivation of the split Bregman method}\label{subsec framework}
Although the split Bregman method originated from Bregman iterations \citep{COSthree:SIAM:2008,CaiOsherShen:MMS:2009,ZBBO:SIIMS:2010}, it has been demonstrated to be equivalent to the alternating direction method of multipliers (ADMM) \citep{GabayMercier:CMA:1976,GlowinskiMarrocco:RFAIRO:1975,Setzer:SSVMCV:2009,ZhangBurgerOsher:JSC:2010}. For simplicity of presentation, next we derive the split Bregman method using the augmented Lagrangian method \citep{Hestenes:JOTA:1969,Rockafellar:MP:1973}.

We first define an augmented Lagrangian function of \eqref{min constraint general likelihood}
\begin{equation}\label{eq Aug Lag}
  \mathcal{L}(\Theta,A,M)=-\log\det\Theta+\hbox{tr}(S\Theta)+
 \lambda\phi(A)+\hbox{tr}(M^T(\Theta-A))+\frac{\mu}{2}\|\Theta-A\|_F^2,
\end{equation}
where matrix $M$ is a dual variable corresponding to the linear constraint $\Theta=A$, and $\mu>0$ is a parameter. Compared with the standard Lagrangian function, the augmented Lagrangian function has an extra term $\frac{\mu}{2}\|\Theta-A\|_F^2$, which penalizes the violation of the linear constraint $\Theta=A$.

With the definition of the augmented Lagrangian function \eqref{eq Aug Lag}, the primal problem \eqref{min constraint general likelihood} is equivalent to
\begin{equation}\label{min primal}
  \min_{\Theta\succ 0,A}\max_{M}\mathcal{L}(\Theta,A,M).
\end{equation}
Exchanging the order of $\min$ and $\max$ in \eqref{min primal} leads to the formulation of the dual problem
\begin{equation}\label{max dual}
  \max_{M}E(M),\quad\hbox{with}\quad E(M)=\min_{\Theta\succ 0,A}\mathcal{L}(\Theta,A,M).
\end{equation}
Note that the gradient $\nabla E(M)$ can be calculated by the following \citep{Bertsekas:book:1982}
\begin{equation}\label{eq gradient}
  \nabla E(M)= \Theta(M)-A(M), \quad\hbox{with}\quad (\Theta(M), A(M))=\arg\min_{\Theta\succ 0, A}\mathcal{L}(\Theta,A,M).
\end{equation}

Applying gradient ascent on the dual problem \eqref{max dual} and using equation \eqref{eq gradient}, we obtain the method of multipliers \citep{Rockafellar:MP:1973} to solve \eqref{min constraint general likelihood}
\begin{equation}\label{eq method of multiplier}
  \begin{cases}
    (\Theta^{k+1},A^{k+1})=\arg\min_{\Theta\succ 0, A}\mathcal{L}(\Theta,A,M^k),\\
    M^{k+1}=M^k+\mu(\Theta^{k+1}-A^{k+1}).
  \end{cases}
\end{equation}
Here we have used $\mu$ as the step size of the gradient ascent. It is easy to see that the efficiency of the iterative algorithm \eqref{eq method of multiplier} largely hinges on whether the first equation of \eqref{eq method of multiplier} can be solved efficiently. Note that the  augmented Lagrangian function $\mathcal{L}(\Theta,A,M^k)$ still contains a nondifferentiable term $\phi(A)$. But different from the original objective function \eqref{min general likelihood}, the function $\phi$ induced nondifferentiable term has now been transferred from terms involving $\Theta$ to terms involving $A$ only. Thus we can solve the first equation of \eqref{eq method of multiplier} through an iterative algorithm that alternates between the minimization of $\Theta$ and $A$,
\begin{equation}\label{eq alternate minimization}
\begin{cases}
\Theta^{k+1}=\arg\min_{\Theta\succ 0} -\log\det\Theta+\hbox{tr}(S\Theta)+
\hbox{tr}((M^k)^T(\Theta-A^k))+\frac{\mu}{2}\|\Theta-A^k\|_F^2,\\
A^{k+1}=\arg\min_A  \lambda\phi(A)+\hbox{tr}((M^k)^T(\Theta^{k+1}-A))+\frac{\mu}{2}\|\Theta^{k+1}-A\|_F^2.
\end{cases}
\end{equation}

The method of multipliers requires that the alternative minimization of $\Theta$ and $A$ in \eqref{eq alternate minimization} be run multiple times until convergence.  However, because  the first equation of \eqref{eq method of multiplier} represents only one step of the overall iteration, it is actually not necessary to be solved completely.  In fact, the split Bregman method (or the alternating direction method of multipliers \citep{GabayMercier:CMA:1976}) uses only one iteration of \eqref{eq method of multiplier}, which leads to the following iterative algorithm for solving \eqref{min constraint general likelihood},
\begin{equation}\label{eq ADMM for general}
\begin{cases}
\Theta^{k+1}=\arg\min_{\Theta\succ 0} -\log\det\Theta+\hbox{tr}(S\Theta)+
\hbox{tr}((M^k)^T(\Theta-A^k))+\frac{\mu}{2}\|\Theta-A^k\|_F^2,\\
A^{k+1}=\arg\min_A  \lambda\phi(A)+\frac{\mu}{2}\|\Theta^{k+1}-A+\mu^{-1}M^k\|_F^2,\\
M^{k+1}=M^k+\mu(\Theta^{k+1}-A^{k+1}).
\end{cases}
\end{equation}

\subsubsection{Convergence Property}
The convergence of the iteration \eqref{eq ADMM for general}  can be derived from the convergence theory of the alternating direction method of multipliers or the convergence theory of the split Bregman method \cite{GabayMercier:CMA:1976,EcksteinDouglas:MP:1992,CaiOsherShen:MMS:2009}.
\begin{theorem}\label{theorem convergence}
Let $\Theta^k$ be generated by \eqref{eq ADMM for general}, and $\Theta^*$ be the unique minimizer of \eqref{min constraint general likelihood}. Then,
\begin{equation*}
    \lim_{k\rightarrow\infty}\|\Theta^k-\Theta^*\|=0.
  \end{equation*}
\end{theorem}
From Theorem \ref{theorem convergence}, the condition for the convergence of the iteration \eqref{eq ADMM for general} is quite mild and even irrelevant to the choice of the parameter $\mu$ in the iteration \eqref{eq ADMM for general}.  This property makes the split Bregman method quite general and easy to implement, which partly explains why the method is gaining popularity recently.

\subsubsection{Updating $\Theta$ and $A$}
We first focus on the computation of the first equation of \eqref{eq ADMM for general}. Taking the derivative of the objective function and setting it to be zero, we get
\begin{equation}\label{eq Theta derivative}
  -\Theta^{-1}+\mu\Theta=\mu A^k-S-M^k.
\end{equation}
It is a quadratic equation where the unknown is a matrix. The complexity for solving this equation is at least $O(p^3)$ because of the inversion involved in \eqref{eq Theta derivative}. Note that because $\phi(\Theta)=\phi(\Theta^T)$, if $\Theta^k$ is symmetric, so is $\mu A^k-S-M^k$.
  It is easy to check that the explicit form for the solution of \eqref{eq Theta derivative} under constraint $\Theta \succ 0$, i.e., $\Theta^{k+1}$, is
\begin{equation}\label{eq update theta}
\Theta^{k+1}=\frac{K^k+\sqrt{(K^k)^2+4\mu I}}{2\mu},\qquad\mbox{where}\quad K^k=\mu A^k-S-M^k.
\end{equation}
Here $\sqrt{C}$ is the square root of a symmetric positive definite matrix $C$. Recall that the square root of a symmetric positive definite matrix $C$ is defined to be the matrix whose eigenvectors are the same as those of $C$ and eigenvalues are the square root of those of $C$. Therefore, to get the update of $\Theta^{k+1}$, one may first compute the eigenvalues and eigenvectors of $K^k$, and then get the eigenvalues of $\Theta^{k+1}$ according to \eqref{eq update theta} by replacing the matrices by the corresponding eigenvalues. This approach is rather slow due to the eigen decomposition step. In the next subsection, we will propose a faster algorithm for solving \eqref{eq Theta derivative} based on Newton's iteration \cite{Higham:BOOK:2008,Higham:MC:1986}.

For the second equation of \eqref{eq ADMM for general}, we have made the data fitting term $\frac{\mu}{2}\|\Theta^{k+1}-A+\mu^{-1}M^k\|_F^2$ separable with respect to the entries of $A$. Thus, if the $\phi(A)$ is separable, then it is very easy to get the solution and the computational complexity would be $O(p^2)$ for most cases. See Section \ref{sec example} for the special case of $\phi(A)=\sum_{i\neq j}|A_{ij}|$. Thus, compared with the update of $\Theta$, the computational time for updating $A$ is minor and mostly negligible. The computation of the third equation of \eqref{eq ADMM for general} is straightforward and the computational cost is also $O(p^2)$, which can be neglected as well. So the overall computational time for running one iteration of  \eqref{eq ADMM for general} mostly depends on how fast $\Theta$ can be updated.

\subsection{Newton's method to calculate the square root of a positive definite matrix}\label{subsec algorithm}
As described above,  the main step to obtain an update of $\Theta^{k+1}$ is calculating the square root of the positive definite matrix $(K^k)^2+4\mu I$, where $K^k$ is a symmetric matrix. One possible way is to calculate all the eigenvalues and eigenvectors of matrix $K^k$, which is computational demanding when the size of the matrix is large.  In this subsection, we will use Newton's method \cite{Higham:MC:1986,Higham:BOOK:2008} to calculate the square root of a positive definite matrix directly instead of using the eigenvalue decomposition of $K^k$. Our numerical experiments show that Newton's method for calculating the square root of a positive definite matrix of the form $K^2+4\mu I$, where $K$ is symmetric, is about four times faster than the one using eigen decomposition.

We begin with finding the positive root of the equation $x^2-a=0 (a>0)$ using Newton's method \citep{Burden:book:2004,CaiOsher:preprint:2010}. That is,
\begin{equation}\label{eq Newton iteration scalar}
x^{k+1}=\frac{1}{2}(x^k+\frac{a}{x^k}).
\end{equation}
The following lemma ensures that \eqref{eq Newton iteration scalar} with $x^0>0$ always converges to $\sqrt{a}$ and the convergence rate is quadratic.
\begin{lemma}\label{lemma convergence of Newton}
If we choose $x^0>0$, then $x^k$ generated by \eqref{eq Newton iteration scalar} is well-defined and satisfies
\begin{equation*}
  |x^{k+1}-\sqrt{a}|\leq \min\left\{\frac{1}{2\sqrt{a}}\big|x^k-\sqrt{a}\big|^2,\frac{1}{2}\big|x^k-\sqrt{a}\big|\right\}.
\end{equation*}
\end{lemma}
\begin{proof}
Since $x^0>0$, $x^1$ is well-defined and $x^1=\frac{1}{2}(x^0+\frac{a}{x^0})\geq\sqrt{a}$.  By induction, $x^k$ is well-defined and $x^k\geq \sqrt{a}$. Moreover, using the iteration \eqref{eq Newton iteration scalar},
\begin{equation*}
  |x^{k+1}-\sqrt{a}|=\left|\frac{1}{2}(x^k+\frac{a}{x^k})-\sqrt{a}\right|=\frac{1}{2|x^k|}|x^k-\sqrt{a}|^2\leq \frac{1}{2\sqrt{a}}|x^k-\sqrt{a}|^2,
\end{equation*}
and
\begin{equation*}
  |x^{k+1}-\sqrt{a}|=\frac{1}{2|x^k|}|x^k-\sqrt{a}|^2=\frac{1}{2}(1-\frac{\sqrt{a}}{x^k})|x^k-\sqrt{a}|\leq \frac{1}{2}|x^k-\sqrt{a}|.
\end{equation*}
\end{proof}

Now we apply Newton's method to calculate the square root of a positive definite matrix of the form $K^2+\alpha I$, where $K$ is symmetric and $\alpha$ is a positive constant. Let $X^0=\sqrt{\alpha}I$ and
\begin{equation}\label{eq Newton iteration Matrix}
  X^{k+1}=\frac{1}{2}\left(X^k+(X^k)^{-1}(K^2+\alpha I)\right).
\end{equation}
Then the iteration \eqref{eq Newton iteration Matrix} always converges quadratically to $\sqrt{K^2+\alpha I}$. Similar algorithms  have been proposed in \citep{Higham:SSSC:1986,HighamRobert:SSSC:1990,CaiOsher:preprint:2010} to compute  polar decompositions.
\begin{theorem}\label{Theorem Newton}
If we choose $X^0=\sqrt{\alpha}I$, then $X^k$ generated by \eqref{eq Newton iteration Matrix} is well-defined and quadratically converges to $\sqrt{K^2+\alpha I}$. More specifically,
\begin{equation}
  \|X^k-\sqrt{K^2+\alpha I}\|_2\leq \min\left\{\frac{1}{2\sqrt{\alpha+\lambda_{\min}(K^2)}}\|X^k-\sqrt{K^2+\alpha I}\|_2^2,\frac{1}{2}\|X^k-\sqrt{K^2+\alpha I}\|_2\right\},
\end{equation}
where $\lambda_{\min}(K^2)$ is the minimum eigenvalue of matrix $K$.
\end{theorem}
\begin{proof}
Let $K=U\Omega U^T$ with $\Omega=\hbox{diag}(\omega_1,\ldots,\omega_p)$ and $UU^T=U^TU=I$. Since $X^0=\sqrt{\alpha}I$, by induction, $X^k$ can be written as $X^k=U\Omega^kU^T$ with $\Omega^k=\hbox{diag}(\omega_1^k,\ldots,\omega_p^k)$ and $\omega_i^k>0$. Furthermore,
$$
X^{k+1}=U\left(\frac{1}{2}\left(\Omega^k+(\Omega^k)^{-1}(\Omega^2+\alpha I)\right)\right)U^T.
$$
Therefore, \eqref{eq Newton iteration Matrix} changes only the singular values which are governed by \eqref{eq Newton iteration scalar}. The Theorem follows immediately from Lemma  \ref{lemma convergence of Newton}.
\end{proof}

Note that the choice of the initial guess $X^0$ can be any matrix of the form $U\Lambda U^T$, where $U$ is the eigenvectors of $K$ and $\Lambda$ is a diagonal matrix with positive diagonal. In the above, we have chosen $X^0=\sqrt{\alpha}I$ for simplicity.

Combining the iteration \eqref{eq ADMM for general} and  Newton's method to calculate the square root of the positive matrix of the form $K^2+4\mu I$ where $K$ is symmetric, we get Algorithm \ref{alg SBGM} to compute the solution of \eqref{min general likelihood}.

\begin{algorithm}[htp]
\caption{Split Bregman method for generalized graphical models \eqref{min general likelihood} (SBGM)}
\label{alg SBGM}
\begin{algorithmic}
\STATE Given $\mu$. Initialize $\Theta^0, A^0,$  and  $M^0$.
\REPEAT
\STATE 1) Compute $K^k=\mu A^k-S-M^k$
\STATE 2) Use Newton method to compute $X^k=\sqrt{(K^k)^2+4\mu I}$
\STATE 3) $\Theta^{k+1}=\frac{K^k+X^k}{2\mu}$
\STATE 4) $A^{k+1}=\arg\min_A  \lambda\phi(A)+\frac{\mu}{2}\|\Theta-A^k+\mu^{-1}M^k\|_F^2$
\STATE 5) $M^{k+1}=M^k+\mu(\Theta^{k+1}-A^{k+1})$
\UNTIL{\STATE Convergence}
\end{algorithmic}
\end{algorithm}

\section{The Split Bregman method for $\ell_1$ penalized graphical models (SBGLasso)}\label{sec example}
In this section, we describe a detailed implementation of the split Bregman method to solve the $\ell_1$ penalized inverse covariance matrix estimation problem  \eqref{min l1 penalized likelihood}. The $\ell_1$ penalty corresponds to a special case of the general penalty used in \eqref{min general likelihood} with $\phi(\Theta)=\sum_{i\neq j}|\Theta_{ij}|$.

Algorithm \ref{alg SBGM} can be applied here with only minor changes. The updates of $\Theta$ and and $M$ are exactly the same as the ones in Algorithm \ref{alg SBGM}. For the update of $A$,  let $\mathcal{T}_\lambda$ be a soft thresholding operator defined on matrix space and satisfying
$$\mathcal{T}_\lambda(\Omega)=(t_\lambda(\omega_{ij}))_{\tiny{\begin{matrix}i,j=1\cr i\neq j\end{matrix}}}^p, \quad \hbox{with}\quad t_\lambda(\omega_{ij})=\hbox{sgn}(\omega_{ij})\max\{0,|\omega_{ij}|-\lambda\}.$$
Then the update of $A$ is
$$A^{k+1}=\mathcal{T}_{\frac{\mu}{\lambda}}(\Theta^{k+1}+\mu^{-1}M^{k}).$$
So we obtain Algorithm \ref{alg SBGLasso} to solve \eqref{min l1 penalized likelihood}.

\begin{algorithm}[htp]
\caption{Split Bregman method for $\ell_1$ penalized graphical models (SBGLasso)}
\label{alg SBGLasso}
\begin{algorithmic}
\STATE Initialize $\Theta^0, A^0,$  and  $M^0$.
\REPEAT
\STATE 1) Compute $K^k=\mu A^k-S-M^k$
\STATE 2) Use Newton method to compute $X^k=\sqrt{(K^k)^2+4\mu I}$
\STATE 3) $\Theta^{k+1}=\frac{K^k+X^k}{2\mu}$
\STATE 4) $A^{k+1}=\mathcal{T}_{\frac{\lambda}{\mu}}(\Theta^{k+1}+\mu^{-1}M^k)$
\STATE 5) $M^{k+1}=M^k+\mu(\Theta^{k+1}-A^{k+1})$
\UNTIL{\STATE Convergence}
\end{algorithmic}
\end{algorithm}

\section{Numerical experiments}\label{sec experiment}
In this section, we use time trials to illustrate the utility of our proposed algorithms.  We first demonstrate the efficiency of the proposed Newton's method for calculating the square root of a positive definite matrix by comparing it with the eigenvalue decomposition method.  We then benchmark the performance of the split Bregman method (SBGLasso) for solving the $\ell_1$ penalized inverse covariance matrix estimation problem, and compare it to the graphical lasso method.  The time trials were all conducted on an Intel Core 2 Duo desktop PC (E7500, 2.93GHz).

\subsection{Newton's method versus eigenvalue decomposition method for computing the square root of a positive definite matrix}
We first illustrate the efficiency of the Newton's method for computing the square root of a positive definite matrix of form $M=K^2+\mu I$, where $K=(B+B')/2$ and $B$ is a $p\times p$ random Gaussian matrix with its entries randomly drawn from the standard Gaussian distribution. The form of the matrix $M$ we choose is mostly motivated by the square root we need to solve in the equation \eqref{eq update theta}. Our algorithm is implemented in Matlab using mex programming. We compare it with the method using matlab function ``schur" on the matrix $K$. More specifically, we use the two-line code ``$[U,D]=schur(K);SR=U*diag(sqrt(diag(D).\hat{~} 2+\mu))*U'$" to compute the square root of $M$.  Note that the upper triangular matrix produced by schur decomposition is exactly diagonal since $K$ is symmetric.

In Table \ref{table Newton method}, we have listed the number of iterations needed for our Newton's method to solve the square root of $M$, where we stop the iterations whenever the relative change of two successive steps is less than $10^{-6}$. We also calculated the relative difference between the result derived by our method and the one by schur decomposition. As shown in Table \ref{table Newton method}, the relative difference  is of precision of order $10^{-9}$, demonstrating that our algorithm is highly accurate. Empirically, we find that our algorithm only needs  $8$ steps to converge and the number of steps is mostly constant as the matrix size increases, which illustrates the quadratic convergence rate of our algorithm given in Theorem \ref{Theorem Newton}. Overall, our algorithm is substantially faster than the schur decomposition based method, saving more than $75\%$ of the computational times on all examples we tested. For example, when $n=2000$, our algorithm takes $7.45$ seconds to find a solution, significantly lower than  the $38.08$ seconds used by  the method based on schur decomposition.

\begin{table}[htp]
\begin{center}
\caption{Experimental results for computing the square root of of $M$. All results are averages of $10$ runs.}\label{table Newton method}
\end{center}
\begin{center}
{\begin{tabular*}{0.75\textwidth}{c||c|c||c||c}
\hline \multirow{2}{*}{p}&\multicolumn{2}{c||}{Newton method}&Schur decomposition&relative\\\cline{2-4}
&iters in Newton method&total time(s)&total time(s)& difference\\\hline
\ $1000$\ &8&$1.20$&$4.54$&$9.69\times 10^{-14}$\\\hline
\ $1500$\ &8&$3.35$&$16.11$&$8.54\times 10^{-13}$\\\hline
\ $2000$\ &8&$7.45$&$38.08$&$3.09\times 10^{-11}$\\\hline
\ $2500$\ &8&$13.56$&$76.03$&$3.75\times 10^{-10}$\\\hline
\ $3000$\ &8&$22.43$&$127.32$&$2.35\times10^{-9}$\\\hline
\end{tabular*}}
\end{center}
\end{table}
\subsection{SBGLasso versus the graphical lasso for $\ell_1$ penalized graphical models}
Next we illustrate the efficiency of the split Bregman method for $\ell_1$ penalized maximum likelihood graphical models (SBGLasso)
using time trials. The graphical lasso proposed by \citet{FHT:Biostatistics:2008} is by far the most efficient method for solving the large-scale inverse covariance matrix estimation problem \eqref{min l1 penalized likelihood}, so we focus our comparison with the graphical lasso method.  SBGLasso is coded in C, linked to a Matlab function, while the graphical lasso is coded in Fortran, linked to an R language function, so it is reasonable to compare them using time trials despite that they are implemented in two different languages.

The stopping criteria of SBGLasso is specified as follows.  Let $\Phi(\Theta^k)=-\log\det\Theta^k+\hbox{tr}(S\Theta^k)+\lambda\sum_{i\neq j}|\Theta_{ij}^k|$. Since the $\ell_1$ penalized maximum likelihood graphical model \eqref{min l1 penalized likelihood} is a special case of model \eqref{min general likelihood}, we have $\lim_{k\rightarrow \infty}\Phi(\Theta^k)=\Phi(\Theta^*)$ by Theorem \ref{theorem convergence}. We terminate the algorithm when the relative change of the energy functional $\Phi(\Theta)$ falls below a certain threshold $\delta$. In addition,  we require the relative difference  $\|\Theta-A\|_F/\|\Theta\|_F$ to be less than $\delta$ to make sure that the resulting solution is also primal feasible. We used $\delta=10^{-4}$ in our simulation. For the graphical lasso, we also set the termination threshold to be $10^{-4}$.

Note that the convergence of Algorithm \ref{alg SBGLasso} is guaranteed no matter what values of $\mu$ are chosen as shown in Theorem \ref{theorem convergence}. However, the choice of $\mu$ can influence the speed of the algorithm since it could affect the number of iterations involved. In our implementation, we found empirically that choosing $\mu=0.5$ works well for the artificial data and $\mu=0.012$ for the gene expression data.

\subsubsection{Artificial data}\label{subsec artificial data}

To generate the artificial data, we first create a set of sparse inverse covariance matrices with different dimension $p$, and then generate $n$ samples from the multivariate Gaussian distribution parametrized by each of the inverse covariance matrices.   The sparse inverse covariance matrices are created using the same procedure as described in \citep{BEd:JMLR:2008}. More specially,  to create an inverse covariance matrix, we first generate a random $p\times p$ diagonal matrix with positive entries, and then symmetrically insert random numbers to approximately $p$ randomly chosen locations of the matrix. Positive definiteness is ensured by adding a multiple of the identity to the matrix if needed.
The penalty parameter is chosen such that the estimated inverse covariance matrix has roughly the same number of nonzero entries as the actual inverse covariance matrix.

Table \ref{table art random} shows a comparison of SBGLasso and the graphical lasso on both computational times as well as relative errors  after testing on the artificial data.  We note that  SBGLasso is significantly faster than the graphical lasso for large-scale data ($p> 500$),  while being able to achieve the same levels of accuracy.  It takes less than half of the computational times used by the graphical lasso for all cases we tested except when $p=500$.  For example, when $n=2000,p=3000$, SBGLasso takes only $269.53$ seconds to find the optimal solution, compared to the $729.63$ seconds used by the graphical lasso.

To evaluate how the performance of SBGLasso scales with the problem size, we plotted the CPU time that SBGLasso took to solve the problem \eqref{min l1 penalized likelihood} as a function of $p$ and $n$. The CPU time is roughly quadratic in $p$ and constant in $n$. This phenomena is expected since each of the concentration matrices used in the artificial data has about $\frac{p(p+1)}{2}$ unknowns , which is quadratic with respect to $p$. By contrast, the number of samples  only appears in $S$, and does not directly affect the speed of SBGLasso.

\begin{table}[htp]
\caption{Comparing the performance of SBGLasso and the graphical lasso on the artificial data}
\begin{center}
{\begin{tabular*}{0.75\textwidth}{ c||c|c||c|c}
\cline{1-5} \multirow{2}{*}{$n$, $p$}&\multicolumn{2}{c||}{SBFGLasso}&\multicolumn{2}{c}{Graphical Lasso}\\\cline{2-5}
&total time (s)&relative error&total time(s)&relative error\\\cline{1-5}
$n=1000,p=500$&$2.61$&$0.1136$&$1.94$&$0.1136$\\\cline{1-5}
$n=1000,p=1000$&$14.09$&$0.1192$&$27.08$&$0.1193$\\\cline{1-5}
$n=1000,p=2000$&$88.71$&$0.1170$&$217.88$&$0.1170$\\\cline{1-5}
$n=2000,p=3000$&$269.53$&$0.0980$&$729.63$&$0.0979$\\\cline{1-5}
\end{tabular*}}
\label{table art random}
\end{center}
\end{table}

\begin{figure}[htp]
\subfigure[]{\includegraphics[width=.48\textwidth]{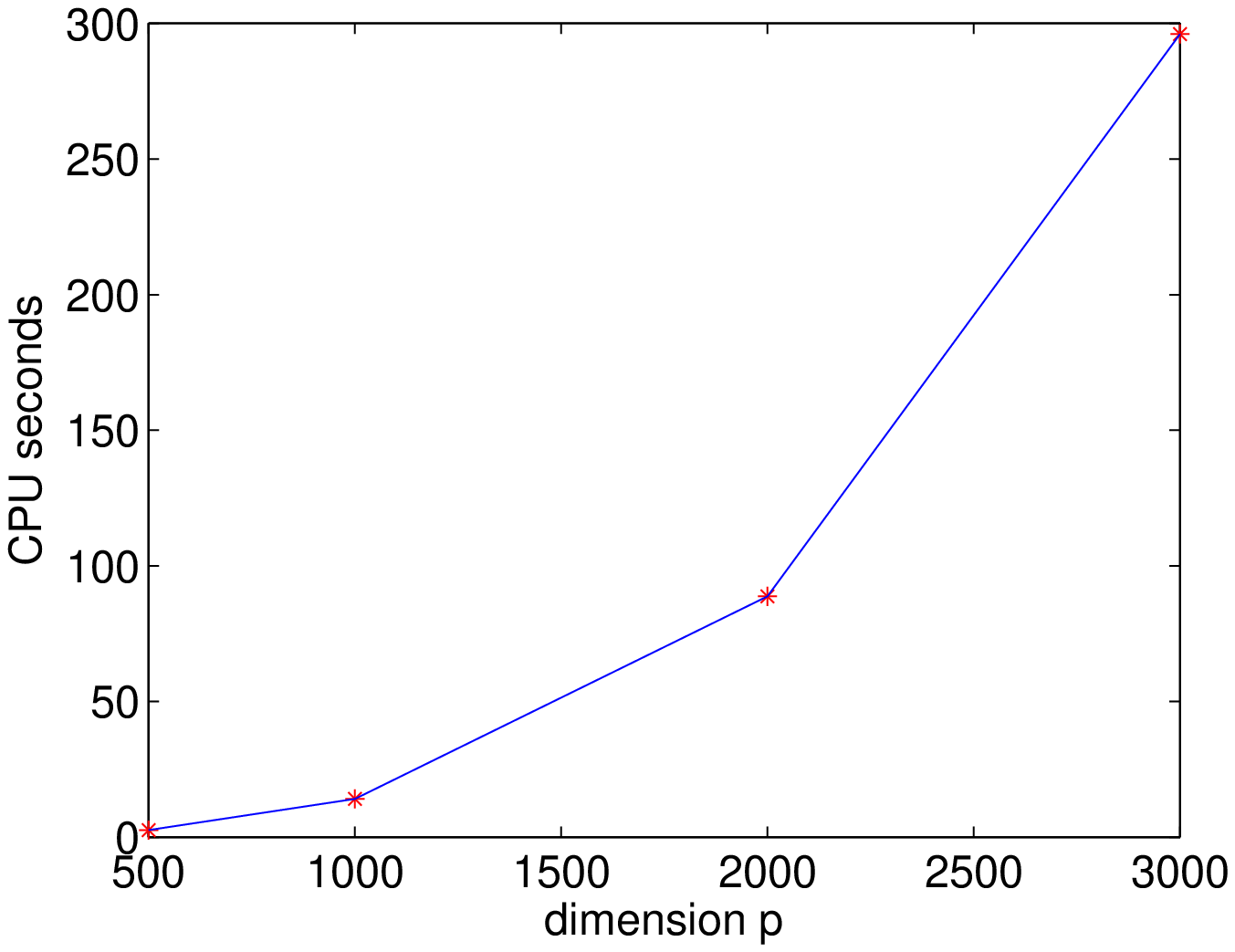}}
\subfigure[]{\includegraphics[width=.48\textwidth]{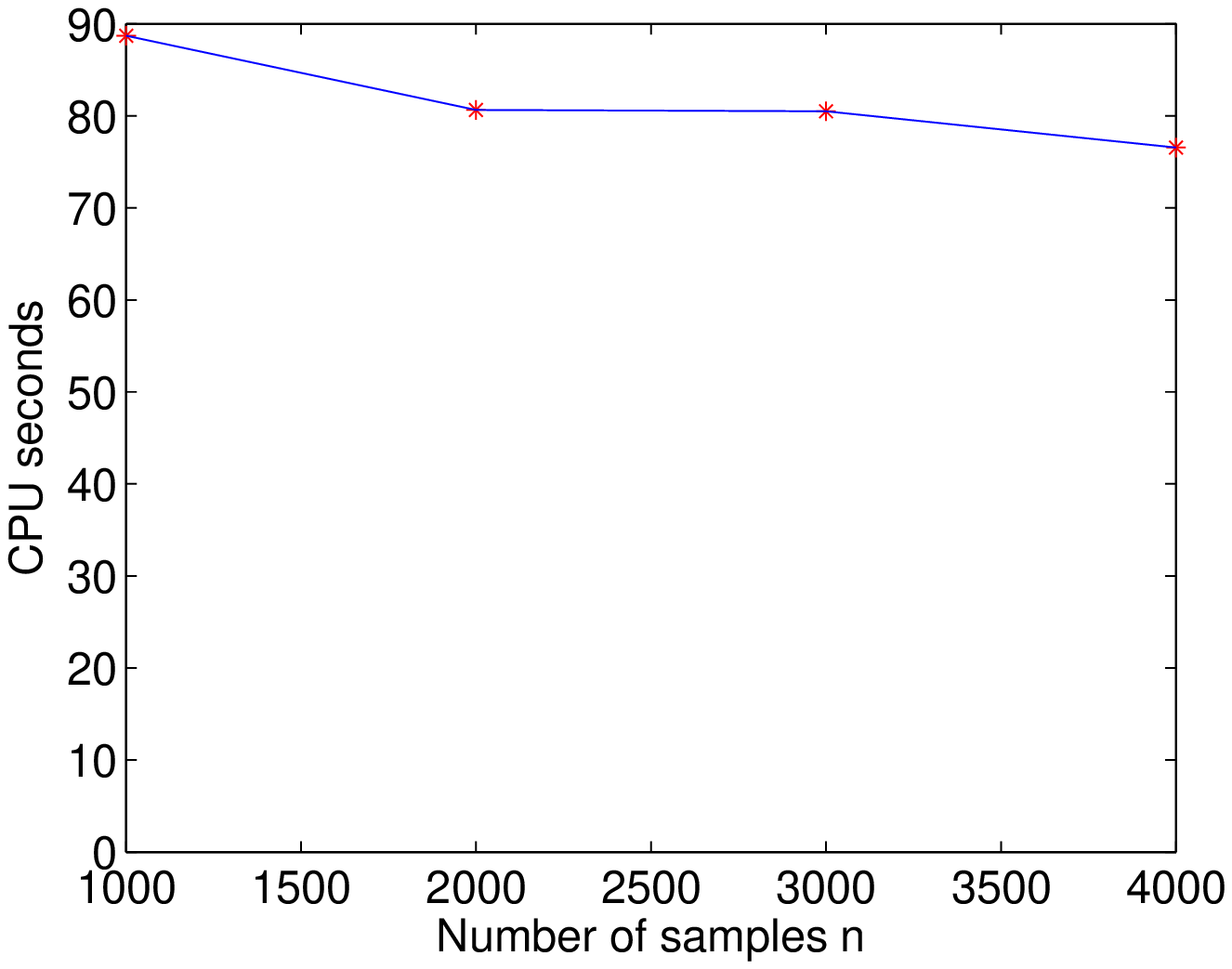}}
\caption{CPU times for SBGLasso for the same problem as in Table \ref{table art random}, for different values of $n$ and $p$. (a) $n$ is fixed and equals to $1000$; (b) $p$ is fixed and equals to $2000$.}\label{figure CPU times}
\end{figure}

\subsubsection{Gene expression data}
We next apply SBGLasso to learn gene regulatory networks from microarray gene expression data. We used the  Rosetta Inpharmatics compendium of gene expression profiles described by \citet{Hughes:Cell:2000}.  The compendium contains microarray measurements of the expressions of 6316 genes in response to 300 diverse mutations and chemical treatments in {\em S. cerevisiae}.  We preprocessed this data by removing genes that either contain missing values in some experiments, or have low variance across different experimental conditions.  This led us to a final dataset of $p=1000$ genes with measurements across $n=300$ experiments.

We used SBGLasso to solve the sparse graphical model \eqref{min l1 penalized likelihood} on this data, and compared its performance to the graphical lasso \citep{FHT:Biostatistics:2008}. The results are summarized in Table \ref{table gene}, which shows the computational times spent by different solvers  for different choices of the regularization parameter $\lambda$. SBGLasso consistently outperforms the graphical lasso,  saving more than  $40\%$ time in  all cases. The concentration matrices derived by SBGLasso and the graphical lasso are very similar. Table \ref{table gene} also lists the values of the objective function achieved when the algorithms converge, showing that the results from the two solvers are almost identical.

\begin{table}[htp]
\caption{Comparing the performance of SBGLasso and the graphical lasso on the gene expression data}
\begin{center}
{\begin{tabular*}{0.75\textwidth}{ c||c|c||c|c}
\cline{1-5} \multirow{2}{*}{$\lambda$}&\multicolumn{2}{c||}{SBFGLasso}&\multicolumn{2}{c}{Graphical Lasso}\\\cline{2-5}
&total time (s)&energy value $\Phi(\Theta)$&total time(s)&energy value $\Phi(\Theta)$\\\cline{1-5}
$\lambda=0.01$&$89.10$&$-2.96\times 10^3$&$167.23$&$-2.96\times 10^3$\\\cline{1-5}
$\lambda=0.015$&$59.95$&$-2.78\times 10^3$&$147.92$&$-2.79\times 10^3$\\\cline{1-5}
$\lambda=0.02$&$54.41$&$-2.72\times 10^3$&$116.18$&$-2.72\times 10^3$\\\cline{1-5}
$\lambda=0.025$&52.94&$-2.69\times 10^3$&$87.74$&$-2.69\times 10^3$\\\cline{1-5}
\end{tabular*}}
\label{table gene}
\end{center}
\end{table}

\section{Discussion}

The problem of inverse covariance matrix estimation arises in many applications.  Because the number of available samples is usually much smaller than the number of free parameters associated with the inverse covariance matrix, a parsimony approach is to select the simplest matrix that adequately explains the data. This leads to the idea of formulating the problem as a regularized maximum likelihood estimation problem with a penalty added to encourage the sparsity of the resulting matrix.  Solving the regularized maximum likelihood problem is however nontrivial for large-scale problems, because of the complexity of the log-likelihood term and the nondifferentiability of the regularization term. In this work, we propose a new approach based on the split Bregman method to solve the regularized inverse covariance matrix estimation problem.  We show that the approach is very fast; it solves a $3000 \times 3000$ matrix estimation problem in less than 5 minutes.

We compared the split Bregman method to the graphical lasso method, the state-of-the-art for solving $\ell_1$ penalized inverse covariance matrix estimation problems  \citep{FHT:Biostatistics:2008}.  We show that our method is about twice faster than the graphical lasso for large-scale problems on both the artificial data and the gene expression data, while being able to achieve the same levels of accuracy.  More importantly, our method is much more general and can handle a variety of sparsity regularization terms other than the $\ell_1$ norm, such as those used in elatistic net, fused lasso and grouped lasso. By contrast, the graphical lasso can only be applied to the $\ell_1$ norm penalty.

The contribution of the paper lies in two aspects. First, we reformulated the regularized maximum likelihood estimation problem by introducing an auxiliary variable, which leads to the decoupling of the log-likelihood term and the regularization term, making the problem amenable to the split Bregman method.  Second, we proposed to use Newton's method to calculate the square root of a positive definite matrix,  the major time consuming step of our algorithm.  Because the matrix quadratic equation is separable in its eigenvalues, the conventional solver is first to compute the eigen decomposition and then solve univariate quadratic equations about the eigenvalues. This approach is generally slow, as the eigen decomposition is time-consuming. To accelerate it, we propose to use a matrix iterative algorithm \cite{Higham:MC:1986,Higham:BOOK:2008} to directly solve the matrix quadratic equation based on Newton's method, without using the eigen decomposition. The contribution is crucial and makes the derived algorithm from the split Bregman method very easy to implement and highly efficient.

\bibliographystyle{unsrtnat}
\bibliography{ygb}
\end{document}